\documentclass{article}
\usepackage{graphicx} 
\usepackage{amsmath, amssymb, amsfonts}
\usepackage{tikz}
\usetikzlibrary{arrows.meta,positioning,fit}
\usepackage{booktabs}
\usepackage{cite}
\usepackage{microtype}              
\usepackage{hyperref}
\usepackage{cleveref}
\title{Semi-Unified Sparse Dictionary Learning with Learnable Top-K LISTA and FISTA Encoders}
\author{Fengsheng Lin,  Shengyi Yan, Trac Duy Tran}
\date{October 2025}

\begin{document}

\maketitle
\begin{abstract}
We present a semi-unified sparse dictionary learning framework that bridges the gap between classical sparse models and modern deep architectures. Specifically, the method integrates strict Top-$K$ LISTA and its convex FISTA-based variant (LISTAConv) into the discriminative LC-KSVD2 model, enabling co-evolution between the sparse encoder and the dictionary under supervised or unsupervised regimes. This unified design retains the interpretability of traditional sparse coding while benefiting from efficient, differentiable training.

We further establish a PALM-style convergence analysis for the convex variant, ensuring theoretical stability under block alternation. Experimentally, our method achieves 95.6\% on CIFAR-10, 86.3\% on CIFAR-100, and 88.5\% on TinyImageNet with faster convergence and lower memory cost ($<$4GB GPU). The results confirm that the proposed LC-KSVD2 + LISTA/LISTAConv pipeline offers an interpretable and computationally efficient alternative for modern deep architectures.
\end{abstract}

\section{Introduction}
\begin{sloppypar}
Sparse representation and dictionary learning have long been studied as interpretable and efficient frameworks for high-dimensional visual modeling. In these paradigms, each sample is expressed as a sparse linear combination of dictionary atoms, capturing meaningful structure and facilitating robust recognition and classification. However, conventional methods such as K-SVD and LC-KSVD rely on fixed or greedy sparse inference algorithms (e.g., OMP or Lasso) that are independent of the dictionary update, which limits optimization efficiency and discriminative coupling. 

Building upon this foundation, our work introduces a semi-unified sparse learning framework that integrates learnable sparse encoders with discriminative dictionary updates. Specifically, we couple strict Top-$K$ LISTA and its convex FISTA-based variant (LISTAConv) within the LC-KSVD2 architecture, allowing the sparse codes and dictionary to co-evolve under both supervised and unsupervised regimes. This joint design retains the interpretability of classical sparse modeling while benefiting from the efficiency and convergence properties of differentiable deep networks.

\end{sloppypar}
\begin{sloppypar}
Importantly, the semi-unified form allows the sparse encoder to adapt to the discriminative structure of each dataset. Through the label-consistency and classifier coupling introduced by LC-KSVD2, the encoder progressively learns to produce sparse representations that are not only more efficient for the given data but also more interpretable, aligning the inference dynamics with the dataset’s intrinsic structure. This adaptive behavior bridges analytical sparse modeling and deep learning practice, enabling the encoder to specialize without sacrificing transparency or stability.
\end{sloppypar}

\begin{sloppypar}
For the convex FISTA-based variant (LISTAConv), we establish a PALM-style convergence analysis, ensuring theoretical stability under convex-block alternation for the \emph{supervised} setting, where the encoder is coupled with discriminative terms
\((A, W)\). In practice, our experiments adopt an \emph{unsupervised} counterpart derived from this formulation by removing the discriminative supervision, yielding a lightweight convex proxy that requires no layer-wise feedback training for \(B\). Despite this simplification, the unsupervised LISTAConv preserves empirical stability and achieves performance highly comparable to the learnable Top-\(K\) LISTA model,
while operating at even lower computational cost. As shown in our results, it converges faster and requires substantially less GPU memory (under \(4\)~GB), with only a slight difference in final accuracy.
\end{sloppypar}
\begin{sloppypar}
Beyond theoretical guarantees, the proposed semi-unified framework demonstrates practical efficiency and interpretability across multiple datasets. On CIFAR-10, CIFAR-100, and TinyImageNet, it consistently improves over the LoRA-tuned baselines while requiring less GPU memory and faster convergence. These results confirm that the LC-KSVD2 + LISTA/LISTAConv pipeline bridges analytical sparse modeling and modern deep architectures, offering a balance between accuracy, stability, and computational efficiency.

\end{sloppypar}
\section{Related Work}
\begin{sloppypar}
Sparse representation and dictionary learning have long served as interpretable tools for image and signal modeling. Early approaches such as \textit{KSVD}~\cite{aharon2006ksvd} learned overcomplete dictionaries that enabled sparse reconstruction of signals. While effective for unsupervised representation, these methods focused solely on reconstruction fidelity and lacked mechanisms for discriminative supervision.
\end{sloppypar}
\begin{sloppypar}
To introduce class awareness, Label-consistent KSVD (\textit{LC-KSVD})~\cite{jiang2013label}  incorporated a label-consistency term and a linear classifier into the dictionary-learning process. By jointly optimizing the dictionary, sparse codes, and classifier, LC-KSVD improved recognition performance on face and object datasets. Subsequent extensions such as the Structured Joint Sparse Discriminative Dictionary Learning model~\cite{jiang2013structured} further enforced structured sparsity across related classes. Nevertheless, these methods rely on alternating optimization with non-differentiable sparse solvers such as Orthogonal Matching Pursuit (OMP) or Lasso, and SVD-based dictionary updates, making them difficult to integrate into modern differentiable pipelines.
\end{sloppypar}
\begin{sloppypar}
In parallel, learnable unrolled encoders emerged to approximate iterative sparse inference within neural-network architectures. \textit{Learned ISTA (LISTA)} framework~\cite{gregor2010learning} unfolded the ISTA algorithm into a sequence of learnable layers and was later shown to possess linear-convergence guarantees in its unfolded variant \textit{unfolded ISTA} model of Chen \textit{et al.}~\cite{chen2018theoretical}. These approaches provide fast and differentiable sparse inference and have inspired a family of deep unrolled optimization networks~\cite{liu2019alista}. \textit{However, these unrolled encoders are typically designed for reconstruction or inverse problems and do not include discriminative supervision such as label consistency.}
\end{sloppypar}
\begin{sloppypar}
A related variant, Label Embedded Dictionary Learning (LEDL)~\cite{shao2019ledl}, replaces the nonconvex $\ell_0$ sparsity regularization in LC-KSVD with a convex $\ell_1$ formulation. This substitution avoids the NP-hardness of $\ell_0$-based sparse coding and allows optimization through the alternating direction method of multipliers (ADMM) and block coordinate descent (BCD) frameworks. While this formulation improves stability and convergence behavior, it remains a fully alternating optimization scheme with separate sparse coding and dictionary updates. In contrast, our framework adopts a semi-unified form that enables gradient-based interaction between the sparse encoder and the dictionary update, allowing partial differentiability and data-driven adaptation while preserving analytical interpretability.
\end{sloppypar}

\section{Proposed Method}
In this section, we present how our method is constructed.
\subsection{Problem Formulation}
For the Top-K Lista version, the minimization takes place over
\begin{equation}
\begin{split}
\mathcal{F}(D,G,A,W)
&= \tfrac{1}{2}\|Y - DG\|_F^2
+ \tfrac{\alpha}{2}\|AG - Q\|_F^2
+ \tfrac{\beta}{2}\|WG - H\|_F^2
+ \tfrac{\varepsilon_D}{2}\|D\|_F^2 \\
&\quad
+ \tfrac{\mu_A}{2}\|A\|_F^2
+ \tfrac{\rho_W}{2}\|W\|_F^2,
\end{split}
\label{eq:lcobj}
\end{equation}
where \( \|g_j\|_0 \le k \) and \( \|d_j\|_2 = 1 \) for every \( j \in \{1,2,\ldots,n\} \),
and the regularization weights are strictly positive:
\( \alpha,\ \beta,\ \mu_A,\ \rho_W,\ \varepsilon_D > 0 \).

For the FISTA case, the objective function of the optimization problem is 
\begin{equation}
\begin{split}
\mathcal{F}(D,G,A,W)
&= \tfrac{1}{2}\|Y - DG\|_F^2
+ \tfrac{\alpha}{2}\|AG - Q\|_F^2
+ \tfrac{\beta}{2}\|WG - H\|_F^2
+ \tfrac{\varepsilon_D}{2}\|D\|_F^2 \\
&\quad
+ \tfrac{\mu_A}{2}\|A\|_F^2
+ \tfrac{\rho_W}{2}\|W\|_F^2
+ \tfrac{\mu_G}{2}\|G\|_F^2
+ \lambda \|G\|_1
+ \iota_{\mathcal{C}}(D),
\end{split}
\end{equation}
where \( \|g_j\|_0 \le k \) and each column of \(D\) is $\ell_2$-normalized
(\( \|d_j\|_2 = 1\) for every \( j \in \{1,2,\ldots,n\}\)),
and the regularization weights are strictly positive:
\( \alpha,\ \beta,\ \mu_A,\ \rho_W,\ \varepsilon_D > 0 \).

To be specific, matrix \(Y\) consists of feature vectors. Here, the \(\mathbb{R}^{K \times N}\) matrix \(Q\) is the label-consistency target matrix defined following LC-KSVD2. 
For each training sample \(y_i\) belonging to class c, the vector \(q_i\) (the i-th column of \(Q\)) 
is constructed such that the entries corresponding to the dictionary atoms associated 
with class \(c\) are set to \(1\) and all other entries are \(0\). This encourages samples from the 
 same class to activate similar subsets of dictionary atoms. The \( \mathbb{R}^{C \times N}\) matrix \(H\) denotes the one-hot label matrix, where each column \(h_j\) indicates the class of sample \(y_j\). The classification term \(\|WG-H\|_F^2\) encourages the learned sparse codes \(G\) to be linearly separable by \(W\). In summary, \(Q\) promotes within-class coherence in the sparse representations, whereas \(H\) enforces between-class separability through the linear classifier \(W\).

The key learning matrices used in our model are summarized below:
\[
\begin{aligned}
& D~\text{(dictionary)}, && G \in \mathbb{R}^{K \times N}~\text{(codes)},\\
& A~\text{(LC matrix)},  && W \in \mathbb{R}^{C \times K}~\text{(classifier)}.
\end{aligned}
\]

The \(\mathbb{R}^{d \times K}\) matrix \(D\) is the learned dictionary containing \(K\) atoms that form the shared representation space. Each category has \(K/C\) atoms.
Real matrix \(G\) denotes the sparse codes of the data in the dictionary space and is obtained through a learnable sparse encoder (Top-K LISTA or LISTAConv/FISTA).
Real matrix \(A\) with size \(K \times K\) is the label-consistency transform that encourages samples of the same class to activate similar atoms by aligning \(AG\) with the target matrix \(Q\).\(W\) is a linear classifier operating on the sparse codes, enforcing between-class separability via the term \(\|WG - H\|_F^2\).
\subsection{Semi-Unified Training}
We consider two forms of sparse inference within the same framework. For the nonconvex model, we use a learnable Top-K LISTA encoder, in which the shrinkage parameters are trained to approximate the optimal sparse codes. For the convex model, we use a deterministic FISTA-based Lasso solver (LISTAConv), which computes \(G\) as the solution to a strongly-convex proxy objective function. The former provides data-adaptive sparse inference, while the latter offers a stable and theoretically convergent encoding process.

In Section~\ref{sec:convergence}, the FISTA-based encoder is analyzed under a \textit{supervised} objective function coupled with (\(A, W\)), ensuring PALM-style convergence under full alternation. However, in practice we found that such supervised coupling may cause the convex encoder to overfit by \(G\) too closely with the training dataset. Therefore, during experiments, we adopted and empirically verified an \textit{unsupervised} FISTA variant that focuses purely on reconstruction, while the Top-\(K\) LISTA branch later fine-tunes its learnable parameters \(\theta\) under discriminative supervision.
\subsubsection{Sparse Encoder Design}
We instantiate the sparse encoder \(G\) in two alternative forms:

\textbf{Learnable Top-$K$ LISTA}

The learnable Top-K LISTA encoder is defined as
\[
G = \text{LISTA}_{\theta}(Y; D),
\]
where \(\theta = B\) denotes the layer-wise feedback matrices. 
The shrinkage operator is a strict Top-\(K\) function \(\mathcal{T}_K(\cdot)\) (with a fixed \(K\) or ratio \(p = T/K\)), 
i.e., thresholds are not learned. A typical layer update is given by
\[
G^{(t+1)} = \mathcal{T}_K\!\big(B^{(t)}Y + (I - B^{(t)}D)G^{(t)}\big),
\]
where \(\mathcal{T}_K(\cdot)\) keeps only the \(K\) largest-magnitude entries per code vector. 
Unlike conventional unsupervised sparse inference, \(B\) is optimized only via the discriminative objective in Eq.~(1) (reconstruction + LC + classification), 
so that gradients flow through \(G\) to update \(B\) while \(K\) (or \(p\)) remains fixed.

\textbf{Deterministic LISTAConv (FISTA-based Lasso)}

The FISTA-based encoder defined as
\[G = \mathop{\arg\min}_G \frac{1}{2} \|y- DG\|^2 + \tfrac{\mu_G}{2}\|G\|^2 +\lambda\|G\|_1\]
is implemented via iterative proximal updates.

This encoder is not trained; it provides a convergent inference process. Section~\ref{sec:convergence} has detailed proofs.

Both encoders operate on the same dictionary \(D\) and produce sparse codes \(G\) in the same representation space, enabling a controlled comparison between 
nonconvex (Top-K) and convex (Lasso) sparse modeling behaviors.
\subsubsection{Dictionary/LC/Classifier Updates}
Given the sparse codes \(G\) obtained from the encoder (either Top-K LISTA or FISTA-based Lasso), the remaining variables \((D,A,W)\) are updated in closed-form or convex subproblems. All updates operate on the same sparse representation \(G\), ensuring that reconstruction, label-consistency, and classification terms are aligned within a shared latent space.
\paragraph{Dictionary Update}
The dictionary \(D\in R^{d\times K}\) is optimized by minimizing the reconstruction objective function:
\[\frac{1}{2}\|Y-DG\|_F^2+\dfrac{\epsilon_D}{2}\|D\|_F^2.\]
where jth column of \(D\)(\(d_j\)) has \(\|d_j\|_2= 1\) 

\subparagraph{Top-K LISTA case}

This objective function is a ridge-regularized least-squares problem and has the closed-form solution
\[D\leftarrow YG^\top (GG^\top+\epsilon_D I)^{-1}.\]
After each update, columns of \(D\) are normalized to unit \(l_2\) to prevent degenerate scaling.

\subparagraph{FISTA case (PGD update)}

Since \(G\) in the FISTA setting comes from a convex sparse inference stage, the dictionary update can be performed by projected gradient descent(PGD)

\[D^{(t+1)} = D^{(t)}- \dfrac{1}{\|GG^\top\|_2+\epsilon_D}((D^\top G-Y)G^\top+\epsilon_D D^{(t)}),\]
followed by column-wise normalization
\[d_j\leftarrow \dfrac{d_j}{\|d_j\|_2}, j=1,...,K.\]
This update avoids computing the matrix inverse and is more scalable when \(K\) is large.

\paragraph{Label-Consistency Matrix and Classifier Update}
The LC matrix \(A \in \mathbb{R}^{K \times K}\) aligns the sparse codes with the class-consistent target matrix \(Q\) by solving
\[
\min_A \frac{\alpha}{2}\|AG - Q\|_F^2 + \frac{\mu_A}{2}\|A\|_F^2,
\]
which admits the closed-form solution:
\[
A \leftarrow QG^\top (GG^\top + \frac{\mu_A}{\alpha} I)^{-1}.
\]
This step encourages samples from the same class to activate similar subsets of dictionary atoms. 
If the regularization weight \(\frac{\alpha}{2}\) is modified, the closed-form expression changes correspondingly. 
A similar update applies to \(W\). 

The linear classifier \(W \in \mathbb{R}^{C \times K}\) is optimized by solving:
\[
\min_W \dfrac{\beta}{2}\|WG - H\|_F^2 + \frac{\rho_W}{2}\|W\|_F^2,
\]
yielding the closed-form solution:
\[
W \leftarrow HG^\top (GG^\top + \dfrac{\rho_W I}{\beta})^{-1}.
\]

\subsection{Two-Stage Training Strategy}
Early supervision risks representation collapse, where the sparse codes G become overly aligned to the training labels and lose generalizable structure. This happens because, in early iterations, the dictionary \(D\), label-consistency mapping \(A\), classifier \(W\), and sparse codes \(G\) are not yet aligned; prematurely enforcing supervision amplifies noise and propagates errors across alternating updates.

To prevent this effect, we employ a warm-up and ramped supervision schedule:
\begin{center}
\vspace{-3pt}
\begin{tikzpicture}[
  font=\bfseries\small,
  box/.style={
    draw,
    rounded corners=0.8pt,
    inner xsep=3pt,     
    inner ysep=0.5pt,   
    text width=3.6cm,   
    align=center
  }
]
\node[box] (left) {Warm-Up Phase\\(unsupervised)};
\node[right=3mm of left] (arrow) {$\Rightarrow$};
\node[box, right=3mm of arrow] (right) {Discriminative Phase\\(supervised)};

\node[
  draw,
  rounded corners=1pt,
  inner sep=1.2pt,  
  fit=(left)(arrow)(right)
] {};
\end{tikzpicture}
\vspace{-4pt}
\end{center}

\subparagraph{Warm-Up Phase (Unsupervised Dictionary Formation)}
During warm-up, the Top-\(K\) LISTA encoder solves a reconstruction-only sparse coding problem:
\[
\min_{D,\,G}\;\frac{1}{2}\|Y - DG\|_F^{2} + \lambda\|G\|_{1}
\quad \text{s.t.}\; \|d_j\|_2=1,\;\; \mathrm{supp}(G_i)\le K.
\label{eq:warmup-topk}
\]
Here, \(\mathrm{supp}(G_i)\le K\) indicates that each code vector \(G_i\) is constrained to have at most \(K\) nonzero entries, which is implemented by a \textbf{strict Top-\(K\) shrinkage} operator in LISTA.

\subparagraph{Discriminative Phase (ramped supervision)}
After warm-up, we gradually introduce label-consistency and classifier terms using a monotone-schedule \(s(t)\in [0,1]\)(linear in our experiments):\[\alpha_t = \alpha_{\max} s(t), \beta_t = \beta_{\max} s(t)\] and update \(G\) with either learnable Top-\(K\) LISTA or deterministic LISTAConv (FISTA-Lasso) solving the convex proxy objective function. \(A\) and \(W\) are closed form(ridge) given \(G\). The ramp avoids an abrupt shift of the coding objective and keeps the supports non-degenerate while \(D, A,\) and \(W\) co-adapt.

\section{Convergence Proof}
\label{sec:convergence}
\section*{Convergence Proof Sketch (PALM Framework)}
We adopt the BKL/PALM convergence framework to analyze the proposed alternating optimization. 
Our conclusion is that, under mild assumptions, 
the entire alternating sequence converges to a \emph{critical point} of the objective function 
(i.e. every limit point is stationary, though not necessarily globally optimal).
This satisfies the desired notion of ``rigorous'' convergence without requiring global rates. 

\subsection*{Notation and Objective Function}

Given training features $ Y \in \mathbb{R}^{d \times N} $, one-hot labels 
\( H \in \mathbb{R}^{C \times N} \), and LC targets 
\( Q \in \mathbb{R}^{K \times N} \),
we optimize over \(D , G, A , W, \) where
\(
D \in \mathbb{R}^{d \times K}  \) is the dictionary, 
\(G \in \mathbb{R}^{K \times N} \) is the codes, 
\(A \in \mathbb{R}^{K \times K}\) is the LC matrix, and \(
W \in \mathbb{R}^{C \times K}  \) is the classifier. We consider the following regularized objective function,
\begin{align}
\mathcal{F}(D,G,A,W)
&= \tfrac{1}{2}\|Y - DG\|_F^2
+ \tfrac{\alpha}{2}\|AG - Q\|_F^2
+ \tfrac{\beta}{2}\|WG - H\|_F^2
+ \tfrac{\varepsilon_D}{2}\|D\|_F^2 \notag\\
&\quad
+ \tfrac{\mu_A}{2}\|A\|_F^2
+ \tfrac{\rho_W}{2}\|W\|_F^2
+ \tfrac{\mu_G}{2}\|G\|_F^2
+ \lambda \|G\|_1
+ \iota_{\mathcal{C}}(D),\label{eq:main_obj}
\end{align}
where \(\mathcal{C}=\{D:\|d_j\|_2=1,\ \forall j\}\), and \(\iota_{\mathcal{C}}\) is an indicator function of the set \(\mathcal{C}\).

\subsection*{Update Scheme}
Each outer iteration alternates over the four blocks:
\[
(D, G, A, W) \quad \longleftarrow \quad 
\text{update each variable with others fixed}.
\]
\begin{enumerate}
    \item {Dictionary update:}
    \[
    D^{t+1} = 
    \arg\min_D \|Y - DG^\top\|_F^2 + \varepsilon_D \|D\|_F^2,
    \quad 
    d_j^{t+1} \leftarrow \frac{d_j^{t+1}}{\|d_j^{t+1}\|_2}.
    \]
    This step is a convex quadratic minimization followed by a projection onto
    the feasible set \( \mathcal{C} = \{ D : \|d_j\|_2 = 1 \} \),
    equivalent to a \emph{Projected Gradient Descent} step.  Since both the objective function and the projection are convex, this is a convex optimization problem, which can be solved by projected gradient descent.
    \item {Sparse code update:}
    \begin{align*}
    G^{t+1} = \arg\min_G
    \tfrac{1}{2}\|Y - D^\top G\|_F^2
    + \tfrac{\alpha}{2}\|A^\top G - Q\|_F^2
    + \tfrac{\beta}{2}\|W^\top G - H\|_F^2\\
    + \lambda \|G\|_1
    + \frac{\mu_G}{2}\|G\|_F^2.
    \end{align*}
    This is a convex problem on \(G\), which can be solved by LISTA or proximal mapping.
    
    \item {Linear transform updates:}
    \[
    A^{t+1} = \arg\min_A \frac{\alpha}{2} \|A G^{t+1} - Q\|_F^2 + \frac{\mu_A}{2} \|A\|_F^2,\]
    \[
    W^{t+1} = \arg\min_W \frac{\beta}{2}\|W G^{t+1} - H\|_F^2 + \frac{\rho_W}{2} \|W\|_F^2,
    \]
    both having closed-form solutions via normal equations
    \(A^{t+1} = (QG^\top)\,(GG^\top+\frac{\mu_A}{\alpha} I)^{-1}\), \(
    W^{t+1} = (HG^\top)\,(GG^\top+\frac{\rho_W}{\beta} I)^{-1}.\)
\end{enumerate}

\subsection*{Convergence Argument (Sketch)}

Under the following mild assumptions:
\begin{enumerate}
    \item Each subproblem is convex and has a unique minimizer.
    \item The projection \( \Pi_{\mathcal{C}}(D) = D / \|d_j\|_2 \) is non-expansive.
    \item The objective function \(\mathcal{F}\) satisfies the \emph{Kurdyka–Łojasiewicz (KL) property}.
\end{enumerate}
Then, by the standard PALM (Bolte–Sabach–Teboulle 2014) argument:
\begin{align*}
& \mathcal{F}(Z^{t+1}) \le \mathcal{F}(Z^\top) - c\|Z^{t+1} - Z^\top\|^2, \\
& \mathcal{F}(Z^\top) \text{ is bounded below}, \\
&  \mathcal{F} \text{ satisfies the KL property,}
\end{align*}
the sequence \( \{Z^\top\} = \{(D^\top, G^\top, A^\top, W^\top)\} \)
converges to a critical point of \(\mathcal{F}\).
Therefore, the entire alternating procedure (dictionary learning + LC + classifier + sparse coding) 
is a \emph{block coordinate descent} scheme under the PALM framework. Each block update is either an exact convex minimization or a projection step. Consequently, the full iteration sequence admits monotonic descent and converges to a stationary point of the nonconvex objective function~\eqref{eq:main_obj}.

\section*{Proofings}
There are four things that we should prove (According to page 12 from BST2014):\\
    (H1) Sufficient decrease: $F(x^{k+1})\leq  F(x^k)- a\|x^{k+1}-x^k\|^2 \quad(a>0)$.\\
    (H2) Relative error: $\exists w^{k+1}\in \partial F(x^{k+1})$ such that $\|w^{k+1}\|\leq b\|x^{k+1}-x^k\|$, where b is a positive number.\\
    3. Boundedness: $D, G, A$ and $W$ are bounded.\\
    4. Kurdyka–Łojasiewicz: $\mathcal{F}$ is a semi-algebraic function.\\

\noindent
\paragraph{PALM-based block convergence proof.}
We adopt a standard \textbf{block-coordinate PALM} proof. 
For each block $i \in \{D, A, W, G\}$, we establish 
(i) a per-block sufficient decrease inequality and 
(ii) a per-block relative-error bound. 
Summing the decreases across one full sweep 
$D \!\rightarrow\! A \!\rightarrow\! W \!\rightarrow\! G$ 
yields the global H1 inequality, 
while stacking the per-block optimality residuals yields the global H2 inequality.

All blocks satisfy the \textbf{PALM assumptions}: 
partial gradients are Lipschitz continuous (with constants $L_i$), 
and steps/penalties $c_i^k > L_i$ are ensured via fixed step sizes or 
monotone backtracking.
The proximal or projection mappings for each block are computable.

In our case, $A$ and $W$ are solved in closed form (yielding zero residuals in H2), 
$D$ is a projected gradient step onto the convex column-norm ball (guaranteeing unique projection), and $G$ is a monotone proximal-gradient (FISTA) step.
Together with boundedness 
(via column-norm constraints and $\ell_2$ regularization) 
and the Kurdyka--Łojasiewicz (KL) property of $\mathcal{F}$, 
these conditions imply that the entire alternating sequence 
$\{z^k\} = \{(D^k, A^k, W^k, G^k)\}$ 
is convergent and approaches a critical point of the nonconvex objective function.

\paragraph{Sufficient decrease and relative error}
In this part, we will prove that our model necessarily decreases. First, with $D$, $A$ and $W$ fixed, the term relevant to $G$ is
\begin{align*}
    f_1(G):=\tfrac{1}{2}\|Y - DG\|_F^2
    + \tfrac{\alpha}{2}\|AG - Q\|_F^2
    + \tfrac{\beta}{2}\|WG - H\|_F^2 
    + \tfrac{\mu_G}{2}\|G\|_F^2
+ \lambda \|G\|_1.
\end{align*}
We separate the smooth and non-smooth part of $f_1(G)$, and define
\begin{align*}
    H_G(G)&:=\tfrac{1}{2}\|Y - DG\|_F^2
    + \tfrac{\alpha}{2}\|AG - Q\|_F^2
    + \tfrac{\beta}{2}\|WG - H\|_F^2 
    + \tfrac{\mu_G}{2}\|G\|_F^2,\\
    f_G(G)&:=\lambda \|G\|_1.
\end{align*}
The gradient of $H_G(G)$ is $-D^{T}(Y-DG) + \alpha A^\top(AG- Q)+ \beta W^\top(WG- H)+ \mu_G G$ and is bounded by $L_G:=\|D^\top D\|_2 + \alpha \|A^\top A\|_2 + \beta \|W^\top W\|_2 + \mu_G$. We say \(L_G\) is the Lipschitz constant for the function \(H_G\). Set the step-size to be $t_k < \frac{1}{L_G}$ and let $c_G^k=\frac{1}{t_k}$. The update for $G$ will be\[G^{k+1} = \operatorname{prox}_{\frac{1}{t_k},f_G}\big(G^k - t_k \nabla H_G(G^k)\big).\]
Now we prove that (H1) holds for \(H_G\) at each step. Denote $G^{k+1}-G^k$ by \(\Delta^k \). By smoothness, we have 
\begin{align}\label{equation:1}
H_G(G^{k+1})\leq H_G(G^k)+ \langle \nabla H_G(G^k), \Delta^k \rangle  +\frac{L_G}{2}\|\Delta^k\|_F^2.\end{align}
Since $G^{k+1}$ is the minimizor of 
\(\langle \nabla H_G(G^k), X- G^k\rangle+\frac{c_G^k}{2}\|X-G^k\|_F^2 + f_G(X),\) 
\begin{align}\label{equation:2}
f_G(G^{k+1})\leq f_G(G_k)-<\nabla H_G(G^k),\Delta^k>- \frac{c_G^k}{2}\|\Delta^k\|_F^2.
\end{align}
Adding up~\eqref{equation:1} and~\eqref{equation:2}, we get
\[H_G(G^{k+1})+ f_G(G^{k+1}) \leq H_G(G^k)+ f_G(G^k)+\frac{1}{2}({L_G}- {c_G^k})\|\Delta^k\|_F^2.\]
By construction, $c_G^k> L_G$, (H1) is satisfied.\\
Next, we prove that (H2) is satisfied at each step. By the first-order optimality condition of the proximal step, there exists $S^{k+1}\in \partial f_G(G^{k+1})$ such that $0 = S^{k+1} + \nabla H_G(G^k)+ c_G^k(\Delta^k)$.
Hence, $S^{k+1}= -\nabla H_G(G^k)-c_G^k\Delta^k.$
Adding $\nabla H_G(G^{k+1})$ on both sides we have
\begin{align}
    S^{k+1}+\nabla H_G(G^{k+1})=\nabla H_G(G^{k+1})-\nabla H_G(G^k)-c_G^k\Delta^k.
\end{align}
Denote \(S^{k+1}+\nabla H_G(G^{k+1})\) by \(W_G^{k+1}\). Note that \( f_G\) and \(H_G\) are convex and \(H_G\) is differentiable, 
\begin{align*}
W_G^{k+1} \in \partial f_1 (G^{k+1}).
\end{align*}
Then
\begin{align*}
\|W_G^{k+1}\|_F&= \|\nabla H_G(G^{k+1})-\nabla H_G(G^k)-c_G^k\Delta^k\|_F \\
&\leq \|\nabla H_G(G^{k+1})-\nabla H_G(G^k)\|_F+\|c_G^k\Delta^k\|_F\\
&\leq(L_G+ c_G^k )\|\Delta^k\|_F,
\end{align*}
suffices to show that (H2) holds for \(f_1\). The first inequality follows from the triangle inequality of the norm, and the second equality is due to the Lipschitz continuity of $\nabla H_G$. We have shown that the block relevant to $G$ has a succinct decrease (H1) and a relative error (H2).

Now we prove that the optimization problem over $D$ satisfies such hypothesis, too. Recall that the term relevant to $D$ is
\begin{align*}
    f_2(D) := \tfrac{1}{2}\|Y - DG\|_F^2+ \tfrac{\varepsilon_D}{2}\|D\|_F^2+ \iota_{\mathcal{C}}(D),
\end{align*}
where \(\mathcal{C}=\{D:\|d_j\|_2=1,\ \forall j\}\). We divide it into smooth and non-smooth parts, and define
\begin{align*}
    H_D(D)&:= \tfrac{1}{2}\|Y - DG\|_F^2+ \tfrac{\varepsilon_D}{2}\|D\|_F^2,\\
    f_D(D)&:= \iota_{\mathcal{C}}(D).
\end{align*}
The first part \(H_D\) is differentiable and its Lipschitz constant is $L_D = \|G\|_2^2+ \varepsilon_D$. Denote $D^{k+1}-D^{k}$ by $\Delta_D^k$. 
$D$ is updated by $D^{k+1} = \operatorname{Proj}_\mathcal{C}(D^{k}-\frac{1}{c_D^k} \nabla H_D(D^k))$, then
\begin{align}\label{Inequality:D_smoothness}
H_D(D^{k+1})&\leq H_D(D^k) +\langle \Delta_D , \nabla H_D(D^{k})\rangle+\frac{L_D}{2}\|\Delta_D\|_F^2
\end{align}
Recall that for any $X\in \mathcal{C}$, we have
\begin{align*}
    \langle D^{k+1}-U, X-D^{k+1}\rangle \geq 0
\end{align*} and $U:=D^k- \frac{1}{c_D^k}\nabla H_D(D^k)$.
Substituting $X= D^k$, we have
\begin{align} \label{Inequality:D_optimalcondition}
    -\| \Delta_D \|_F^2 -  \frac{1}{c_D^k} \langle \Delta_D, \nabla H_D(D^k)\rangle \ge 0,
\end{align}
By~\eqref{Inequality:D_smoothness} and~\eqref{Inequality:D_optimalcondition},
\begin{align*}
H_D(D^{k+1})\leq H_D(D^{k})-(c_D^k-\frac{L_D}{2})\|\Delta_D\|_F^2
\end{align*}
When $c_D^k>\frac{L_D}{2}$, (H1) passes.\\
Next, we prove that (H2) is satisfied. Due to Projected first-order optimality condition, there exists $v_D^{k+1}\in N_{C_D}(D^{k+1})$ such that
\begin{align}\label{Equation:D_optimalcondition}
    0 = v_D^{k+1} + \nabla H_D(D^k) + c_D^k \Delta_D.
\end{align}
Adding up the gradient of \(H_D\) and a subgradient of \(\iota_D\) at \(D^{k+1}\), we have $W_D^{k+1}:=v_D^{k+1}+ \nabla H_D(D^{k+1})\in \partial f_2 (D^{(k+1)})$. From~\eqref{Equation:D_optimalcondition}, \(v_D^{k+1} = -\nabla H_D(D^k) - c_D^k \Delta_D\), so
\begin{align*}
    \|W_D^{k+1}\|_F & = \|\nabla H_D(D^{k+1})- \nabla H_D(D^k) - c_D^k \Delta_D\|_F\\&\leq \|\nabla H_D(D^{k+1})- \nabla H_D(D^k)\|_F + c_D^k\|\Delta_D\|_F\\&\leq (L_D+c_D^k)\|\Delta_D\|_F.
    \end{align*}
The first inequality comes from the triangle inequality and the second inequality follows from the Lipschitz continuity
of \(\nabla H_D\). Therefore, it is proved that (H2) holds for $f_2$.\\
The last part of the section proves that the hypothesis (H1) and (H2) hold for the optimization subproblems of $A$ and $W$. Since they are updated in the same way by ridge closed form, here we only present the proof with respect to $A$ without loss of generality. Recall that $A$ is updated by solving the problem as follows
\begin{align*}
    \min_A f_3(A):= \frac{\alpha}{2}\|AG-Q\|_F^2+\frac{\mu_A}{2}\|A\|_F^2,
\end{align*}
where $\mu_A>0$.\\
Let $A$ be in the ridge closed-form that
\begin{align*}
    A^{k+1}= QG^\top(GG^\top +\frac{\mu_A}{\alpha}I)^{-1}.
\end{align*}
Denote the second-differential of \(f_3\) by \(m_A\). Since \(m_A = \mu_A + \alpha\lambda_{min}(GG^\top)\geq \mu_A > 0 \), the objective function is strongly-convex. Then \(\|A^{k+1}\|\) is a global minimizer of \(f_3\),  and
\begin{align*}
    f_3(A^k)-f_3(A^{k+1})\geq \frac{m_A}{2}\|A^{k+1}-A^k\|_F^2,
\end{align*}
which shows that (H1) holds for \(f_3(A)\). To prove that (H2) holds as well, we gain from the optimality that $\nabla f_3(A^{k+1})=0$. Immediately, we get \(W_A^{k+1}: = \nabla f_3(A^{k+1}) \leq \rho_A \|A^{k+1}-A^k\|\) for any $\rho_A>0 $.

\paragraph{Boundedness}
In this section, we prove that $D, A, W$ and $G$ are bounded. Note that $D$ is normalized at each step, $\|D\|_F\leq 1$. We now argue that \(G\) is bounded. Because of the convexity of $f_1(G)$, there exists an optimal solution $G^{*}$ with \(f_3(G^*) \leq f_3 (\mathbf{0})\), i.e.,
\begin{align*}
    \tfrac12\|Y-DG^{*}\|_F^2+\tfrac\alpha2\|AG^{*}-Q\|_F^2+\tfrac\beta2\|WG^{*}-H\|_F^2 +\frac{\mu_G}{2}\|G^{*}\|_F^2 + \mu \|G^{*}\|_1 \\ \leq \frac{1}{2}(\|Y\|_F^2 + \alpha \|Q\|_F^2 + \beta \|H\|_F^2)
\end{align*}
Each term on the left side of the inequality is nonnegative. Hence, we get $\|G^*\|_F^2\leq \frac{1}{\mu_G}(\|Y\|_F^2 + \alpha \|Q\|_F^2 + \beta \|H\|_F^2)$ and $\|G^*\|\leq \frac{1}{\mu}(\|Y\|_F^2 + \alpha \|Q\|_F^2 + \beta \|H\|_F^2)$. 
Next, we prove the boundedness of $A$ and $W$. 
Recall the ridge closed form of $A$ and $W$ as follows
\begin{align*}
    A^*&= QG^\top(GG^\top+ \frac{\mu_A}{\alpha} I)^{-1}\\
    W^*&=  HG^\top\,(GG^\top+\frac{\rho_W}{\beta} I)^{-1}
\end{align*}
Using Singular Value Decomposition, we can have unitary matrices $U, V$ and diagonal matrix $\Sigma$ such that $G^\top = U\Sigma V$. Then, $\|QG^\top\|_F = \|QU\Sigma V\|_F = \|QU\Sigma\|_F\leq \|QU\|_F\|\Sigma\|_2 = \|Q\|_F\|\Sigma\|_2$.
The equality follows from the fact that the Frobenius norm is invariant under unitary transformations. The inequality comes from \(\|AB\|_F \leq \|A\|_F \|B\|_2\) for all \(A, B\). Meanwhile, $GG^\top$ is positive semi-definite and, therefore, Hermitian. By Weyl's inequality, we have
\begin{align*}
    \lambda_{min}(GG^\top + \mu_A I) \geq \mu_A
\end{align*}
Hence,
\begin{align*}
\|A^*\|_F 
&= \Big\| Q G^\top \big(GG^\top + \frac{\mu_A}{\alpha} I \big)^{-1} \Big\|_F \\
&\le \|Q G^\top\|_F \, \Big\| \big(GG^\top + \frac{\mu_A}{\alpha} I \big)^{-1} \Big\|_2 \\
&= \frac{1}{\lambda_{\min}\big(GG^\top + \frac{\mu_A}{\alpha} I \big)} \, \|Q G^\top\|_F \\
&\le \frac{\alpha}{\mu_A} \, \|Q G^\top\|_F \\
&\le \frac{\alpha}{\mu_A} \, \|Q\|_F \, \|G\|_2
\end{align*}
Therefore, we have $A$ bounded because $\|G\|_2 \leq \|G\|_F$ and we have shown that \(\|G\|_F\) is bounded. We omit the proof of the boundedness of \(W\), since it can be established using an approach similar to the one we used for \(A\).
\paragraph{KL property of $F$.} 
Consider
\[
\begin{aligned}
F(D,G,A,W)
&=\tfrac12\|Y-DG\|_F^2+\tfrac\alpha2\|AG-Q\|_F^2+\tfrac\beta2\|WG-H\|_F^2 \\
&\quad+\tfrac{\varepsilon_D}{2}\|D\|_F^2+\tfrac{\mu_A}{2}\|A\|_F^2+\tfrac{\rho_W}{2}\|W\|_F^2+\tfrac{\mu_G}{2}\|G\|_F^2 \\
&\quad+\lambda\|G\|_1+\iota_{\mathcal C}(D),\qquad
\mathcal C=\{D:\ \|d_j\|_2=1,\ \forall j\}.
\end{aligned}
\]
Each term is semi-algebraic: quadratic/Frobenius-norm terms are polynomials; the
$\ell_1$-term is semi-algebraic; the constraint set $\mathcal C$ is a finite Cartesian
product of spheres $\{\|d_j\|_2^2=1\}$, hence a closed semi-algebraic set, so its indicator
$\iota_{\mathcal C}$ is semi-algebraic. Since the class of semi-algebraic functions is
closed under finite sums and linear mappings, $F$ is semi-algebraic. Moreover, $F$ is
proper and lower semi-continuous. Therefore, by the Kurdyka–\L ojasiewicz theorem for
semi-algebraic functions, $F$ satisfies the KL property at every point of its domain.

\section{Experimental Setup}
We evaluate our method on CIFAR-10, CIFAR-100 and TinyImageNet, following standard train/validation splits. All models use features extracted from a frozen backbone unless otherwise stated. For Cifar100 and TinyImageNet, we apply a one-time LoRA fine-tuning step to ViT-B/16 for 8 epochs(rank \(r=8,\alpha=16,droupout = 0.05\)), which requires approximately 45 minutes and \(10.9 \)GB of peak GPU memory. After this stage, the backbone remains frozen, and our method operates only on extracted features.

Given training features \(Y\in R^{d\times N}\) and labels \(y\), the dictionary learning component employs a dictionary \(D\in R^{d\times K}\), sparse codes \(G\in R^{K\times N}\), a label-consistency matrix \(A\in R^{K\times K}\), and a linear classifier \(W\in R^{C\times K}\). We consider twp sparse encoding variants:
\begin{itemize}\setlength\itemsep{0.2em}
    \item Top-K LISTA (nonconvex strict sparsity, learnable shrinkage matrices)
    \item FISTA-based LISTAConv (convex Lasso encoder, no trainable encoder parameters)
\end{itemize}

Unless otherwise noted, we use:
\begin{itemize}\setlength\itemsep{0.2em}
    \item Dictionary size \(K = 900\)(TinyImageNet), \(K=520\)(CIFAR-10/100)
    \item Strict Top-K ratio \(T/K\approx 0.1\) or convex \(\lambda = 1e-2\)
    \item Warm-up of \(2\) outer iterations, followed by \(3\) ramp iterations
    \item \(30\) maximum outer iterations total(converges in \(6\) epochs(\(<5\) minutes) on a single GPU)
\end{itemize}
While both CIFAR-10 and CIFAR-100 use \(K=520\), the choice of \(T\) influences the
balance between class-specific sparsity and cross-class feature sharing.
In LC-KSVD2, dictionary atoms tend to form class-subspace clusters of size
approximately \(K/C\). A larger \(T\) allows samples to additionally activate 
atoms from visually related classes, improving robustness when class affinity
is beneficial, whereas a smaller \(T\) enforces tighter discriminative separation.

For CIFAR-10 (\(C=10\)), inter-class variation is large, so we use \(T=50\) to allow
moderate feature sharing without harming separability. For CIFAR-100 (\(C=100\)),
many classes are semantically and visually similar; thus we use a smaller 
\(T=20\) to enforce stronger class-subspace compactness. For TinyImageNet, the 
larger dictionary size \(K=900\) would suggest \(T\approx 90\) in the same ratio, 
but increasing \(T\) results in nearly linear growth in peak GPU memory. We 
therefore set \(T=10\) to keep refinement lightweight (\(<5\)GB) while preserving
effective sparsity control.

The convex variant(FISTA) updates only \(D, A, W\), yielding \(\approx 1.68 M \) parameters(TinyImageNet). The Top-\(K\) variant additionally learns LISTA shrinkage matrices, totaling \(\approx 6.54M\) parameters. Both are significantly smaller than transformer backbones(typically \(50M-300M+\) parameters)~\cite{dagli2023astroformer}.  When training ends, trained dictionary \(D\) is used to obtain a sparse encode \(G_{test}\) via \(G_{test} = \mathrm{LISTA}_{\theta}(Y_{test};D)\) or let fista solve the problem. And then we classify using \(WG_{test}\).

\section{Results and Analysis}
\subsection{TinyImageNet}
We follow the standard TinyImageNet train/valid split and extract features using a frozen ViT-B/\(16\) backbone. As noted in §4, we optionally apply a one-time LoRA fine-tuning step to the backbone, which takes approximately \(45\) minutes and \(10.9\)GB peak GPU memory. After this, the backbone remains frozen and all learning takes place in the dictionary learning and sparse inference stage.

We evaluate two variants of our refinement stage:
\begin{itemize}\setlength\itemsep{0.2em}
    \item LCKSVD2+ Top-K LISTA
    \item LC-KSVD2+ LISTAConv
\end{itemize}
The comparison against prior methods without using extra training data is shown below:
\begin{table}[h]
\centering
\begin{tabular}{l c}
\toprule
\textbf{Method} & \textbf{Top-1 Accuracy } \\
\midrule
PreActResNet (Ramé et al., 2021)             & 70.24 \\
ResNeXt-50 (SAMix+DM) (Liu et al., 2022)     & 72.39 \\
Context-Aware Pipeline (Yao et al., 2021)    & 73.60 \\
WaveMixLite (Jeevan et al., 2023)            & 77.47 \\
DeiT (Lutati \& Wolf, 2022)                  & 92.00 \\
\textbf{Astroformer (Tseng et al., 2023)}    & \textbf{92.98} \\
\textbf{Ours (LC-KSVD2 + Top-K LISTA)}       & \textbf{88.54} \\
\textbf{Ours (LC-KSVD2 + LISTAConv)}         & \textbf{88.40} \\
\bottomrule
\end{tabular}
\caption{Tiny ImageNet classification accuracy without extra training data.}
\label{tab:tinyimagenet_results}
\end{table}

Our refinement stage improves the pretrained backbone representation in about {3--4} minutes and under 4 GB peak memory, while introducing significantly fewer trainable parameters than transformer-based baselines.
\subsection{CIFAR-10 \& CIFAR-100}
\paragraph{CIFAR-10(no LoRA).}
We use frozen Vit-B/\(16\) features and apply both encoders with \(K= 520, T=50\) Top-\(K\) LISTA reaches \(95.6\%\) in \(\sim 17\) iterations, while the LISTAConv(FISTA-based) reaches \(94.65\%\) in about {6--7} outer iterations. 
\begin{table}[h]
\centering
\begin{tabular}{l c c}
\toprule
\textbf{Method} & \textbf{Iters to conv.} & \textbf{Top-1 Acc.} \\
\midrule
Ours (LC-KSVD2 + LISTAConv)  & $\sim${6--7} outer iters & 94.65\% \\
Ours (LC-KSVD2 + Top-K LISTA) & $\sim$17 outer iters  & 95.60\% \\
\bottomrule
\end{tabular}
\caption{CIFAR-10 with frozen ViT-B/16 features ($K\!=\!520,\;T\!=\!50$).}
\end{table}

\begin{figure}[t]
    \centering
    \includegraphics[width=0.55\linewidth]{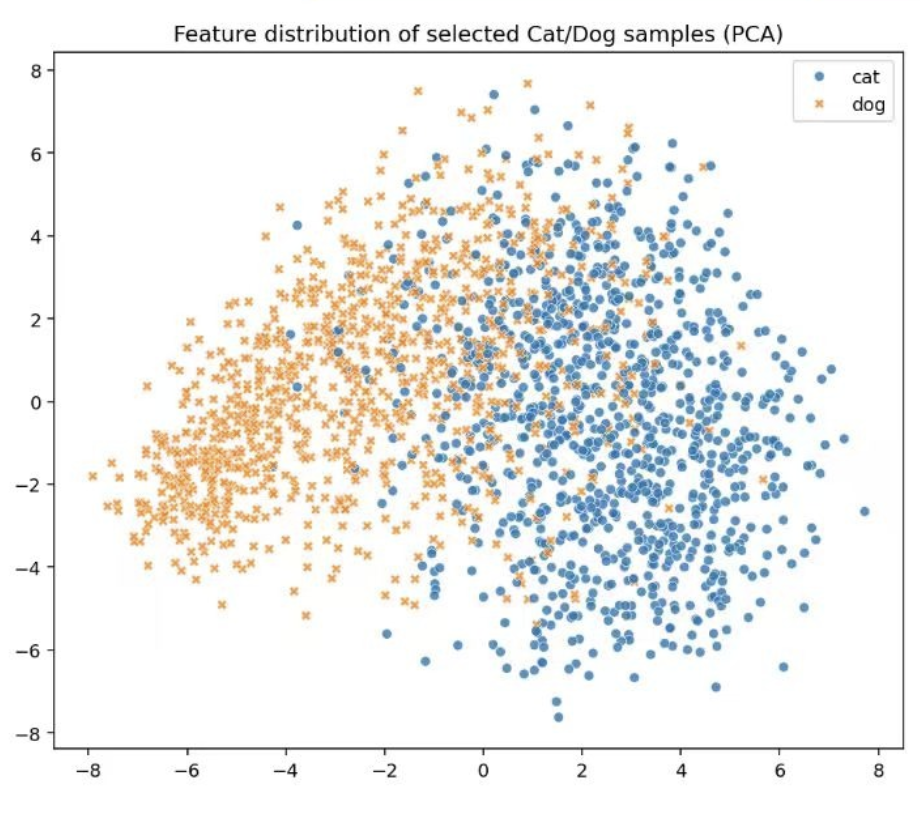}
    \caption{PCA visualization of cat vs. dog samples using frozen ResNet-50 
    features. The two classes remain highly entangled, and sparse refinement 
    alone cannot yield stable separation.}
    \label{fig:catdog_pca}
\end{figure}

\paragraph{Cat–Dog feature overlap on  CIFAR10(ResNet-50)}
At first when we tested performance on CIFAR10, we examined frozen ResNet-50 extracted features. As shown in Figure
~\ref{fig:catdog_pca}, cat and dog samples occupy heavily entangled regions in the representation space. In this case, Top-\(K\) LISTA exhibits limited improvement among cats and dogs: the combined classification error for the two classes remains approximately constant (e.g. cat\(\approx 15\%, dog \approx 15\%, or cat\approx 20\%, dog \approx 10\%\)). This indicates that the backbone feature geometry itself lacks the separability
required to disentangle the two clusters, and post-hoc sparse refinement alone
cannot resolve this overlap.
\paragraph{CIFAR-100 (one-time LoRA)}
We first run a one-time LoRA finetune on ViT-B/\(16\); this baseline achieves \(85.11\%\). Using the same features, we refine with our sparse encoders using \(K = 520, T = 20 \). LISTAConv(convex) attains \(85.15\%\) and converges in \(\sim 4\) outer iterations (\(\sim 1\) min on a single GPU), while TOp-\(K\) LISTA attains \(86.3\%\) and typically stabilizes in \(\sim 15\) outer iterations.
\begin{table}[h]
\centering
\begin{tabular}{l c c c}
\toprule
\textbf{Method} & \textbf{Time / iters} & \textbf{Top-1 Acc.} \\
\midrule
LoRA-tuned ViT-B/16 (baseline) &  8 epochs (one-time) & 85.11\% \\
Ours (LC-KSVD2 + LISTAConv)     & $\sim$4 outer iters  & 85.15\% \\
Ours (LC-KSVD2 + Top-K LISTA)    & $\sim$15 outer iters & 86.3\% \\
\bottomrule
\end{tabular}
\caption{CIFAR-100 with ViT-B/16 + one-time LoRA; refinement uses $K\!=\!520,\;T\!=\!20$.}
\end{table}

\bibliographystyle{ieeetr}
\bibliography{references}  
\end{document}